\documentclass[lettersize,journal]{IEEEtran}
\usepackage{amsmath,amsfonts}
\usepackage{algorithmic}
\usepackage{algorithm}
\usepackage{array}
\usepackage[caption=false,font=normalsize,labelfont=sf,textfont=sf]{subfig}
\usepackage{textcomp}
\usepackage{stfloats}
\usepackage{url}
\usepackage{verbatim}
\usepackage{graphicx}
\usepackage{lettrine}
\usepackage{cite}

\usepackage{hyperref}
\usepackage{cleveref}
\usepackage{float}
\usepackage{booktabs}
\usepackage{multirow} 
\usepackage{xcolor} 
\usepackage{colortbl}
\usepackage{tabu}
\begin{document}

% \title{Bridging the Geometric Gap in Cross-View Geo-Localization via Efficient Adaptation of Vision Foundation Models}

%\title{Bridging the Geometric Gap in Cross-View Geo-Localization via Efficient Adaptation of the Vision Foundation Model DINOv3}

%\title{BGG: Efficient Adaptation from the Vision Foundation Model for Cross-View Geo-Localization by Bridging the Geometric Gap Across Images}

%BGG: Vision Foundation Model for Cross-View Geo-Localization by Bridging the Geometric Gap Across Images

%\title{BGG: Bridging the Geometric Gap between Cross-View images by an Efficient Adaptation for Geo-Localization}

\title{BGG: Bridging the Geometric Gap between Cross-View images by Vision Foundation Model Adaptation for Geo-Localization}

\author{Wei Wang, Dou Quan,~\IEEEmembership{Member,~IEEE}, Ning Huyan,~\IEEEmembership{Member,~IEEE}, Shuang Wang,~\IEEEmembership{Senior Member,~IEEE}, Yi Li,~\IEEEmembership{Member,~IEEE}, Pei He,~\IEEEmembership{Member,~IEEE}, Licheng Jiao,~\IEEEmembership{Life Fellow, IEEE}
		% \thanks{Manuscript created January 2026; This work was supported in part by the National Natural Science Foundation of China under Grant 62201407, 62501356, and 62271377; in part by the Key Research and Development Program of Shannxi Program under Grant 2023QCYLL28, Grant 2024GX-ZDCYL-02-08, and Grant 2024GX-ZDCYL-02-17; in part by the China Postdoctoral Science Foundation under Grant 2022M722496; and in part by the Key Scientific Technological Innovation Research Project by Ministry of Education. (Corresponding author: Dou Quan, Ning Huyan, e-mail: dquan@stu.xidian.edu.cn; quandou@xidian.edu.cn; n-hy@mail.tsinghua.edu.cn.)}
		\thanks{Wei Wang, Dou Quan, Shuang Wang, Pei He, and Licheng Jiao are with the Key Laboratory of Intelligent Perception and Image Understanding of Ministry of Education of China, Xidian University, Yi Li is with the School of Telecommunications, Xidian University, Xi'an 710071, China. Ning Huyan is with the Department of Automation, Tsinghua University, Beijing 100084, China.}}
	\markboth{}%
	{BGG: Bridging the Geometric Gap between Cross-View images by Vision Foundation Model Adaptation for Geo-Localization}
\maketitle

\begin{abstract}
Geometric differences between cross-view images, such as drone and satellite views, significantly increase the challenge of Cross-View Geo-Localization (CVGL), which aims to acquire the geolocation of images by image retrieval. To further enhance the CVGL performance, this paper proposes a parameter-efficient adaptation framework for bridging the geometric gap across images based on the vision foundation model (VFM) (e.g., DINOv3), termed \textbf{BGG}. BGG not only effectively leverages the general visual representations of VFM and captures the robust and consistent features from cross-view images, but also utilizes the generalization capabilities of the VFM, significantly improving the CVGL performance. It mainly contains a Multi-granularity Feature Enhancement Adapter (MFEA) and a Frequency-Aware Structural Aggregation (FASA) module. Specifically, MFEA enhances the scale adaptability and viewpoint robustness of features by multi-level dilated convolutions, effectively bridging the cross-view geometric gap with small training costs. Additionally, considering the \texttt{[CLS]} token lacks spatial details for precise image retrieval and localization, the FASA module modulates patch tokens in the frequency domain and performs adaptive aggregation for local structural feature enhancement. Finally, BGG fuses the enhanced local features with the \texttt{[CLS]} token for more accurate CVGL. Extensive experiments on University-1652 and SUES-200 datasets demonstrate that BGG has significant advantages over other methods and achieves state-of-the-art localization performance with low training costs. The code will be released at \href{https://github.com/Miraitowa515/BGG}{https://github.com/Miraitowa515/BGG}. 

\end{abstract}


\begin{figure}[tbp]  
    \centering
    \includegraphics[width=1\linewidth]{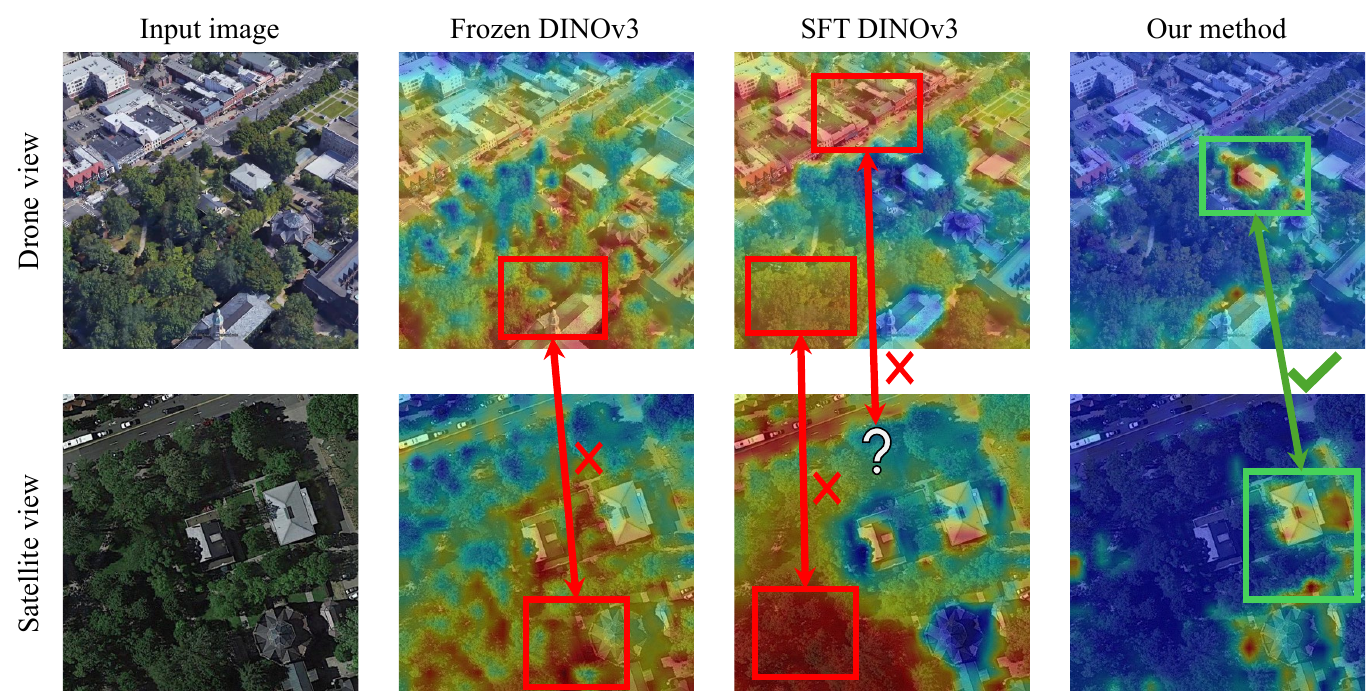}
    \caption{The feature maps of cross-view images captured by the Frozen DINOv3, full-parameter supervised fine-tuning (SFT) DINOv3, and our proposed method. Green and red bounding boxes denote consistent and inconsistent attended regions across views, respectively. The Frozen and SFT models tend to focus on background regions (e.g., tree textures) and struggle to maintain consistent attention on objects between cross-view images. In contrast, our method focuses on discriminative structural regions (e.g., building edges) and exhibits high consistency across images.}
   \label{fig:introductionmelf}
\end{figure}

\section{Introduction}
\lettrine{C}{ross-View} Geo-Localization (CVGL) aims to infer the geometric location of images by image matching and retrieval, which are captured from different viewpoints, such as satellite view and ground view \cite{early_1, CVM_net}, or drone view and ground view \cite{MCCG}. CVGL is crucial for various practical applications \cite{tnnls_retrieval, University, tnnls_application_1}, such as autonomous driving, robotic navigation, object detection, and urban surveillance, particularly in Global Navigation Satellite System (GNSS)-denied environments \cite{GNSS_limit}. However, the drastic geometric variations between cross-view images, including spatial layout discrepancies, scale changes, and geometric deformations, increase the challenges of CVGL. 

Most existing deep learning-based methods use convolutional neural networks (CNNs) to extract features for cross-view image matching. However, due to the limitation of local receptive fields of CNNs, some studies introduce transformer-based architectures \cite{L2LTR, Transgeo} or hybrid networks \cite{SAID, MCCG} to capture long-range dependencies and global contextual information. These methods typically use a high-dimensional \texttt{[CLS]} token as a global descriptor for matching, or employ pooling and aggregation mechanisms (e.g., NetVLAD \cite{NetVLAD}, GeM \cite{GeM}) over local patch features. In addition, some studies attempt to optimize the CVGL model from the perspectives of hard negative mining \cite{Sample4geo} and data augmentation \cite{LCM}.

To deal with the significant viewpoint differences and geometric variations between cross-view images, existing CVGL methods primarily focus on extracting viewpoint-invariant features for matching \cite{SDPL, Game4loc}. They explicitly establish cross-view correspondences through geometric and projection transformations \cite{PCL, SAFA}, or implicitly align the feature distributions across views using domain adaptation techniques \cite{F3_Net, DAC}. Additionally, some approaches focus on fine-grained feature mining \cite{LPN, SRLN} and invariant feature learning \cite{RK_net}. 

Although existing deep learning methods have achieved notable progress, they usually have complex network architectures with high training costs and exhibit limited generalization capability in large-scale or highly dynamic application scenarios. In recent years, visual foundation models (VFMs) (e.g., DINOv3 \cite{Dinov3}) have achieved remarkable success in learning general visual feature representations and strong generalization \cite{PETL, tnnls_prompt_1}. Nevertheless, few studies have introduced the powerful VFM into the CVGL task. 

Due to the severe geometric variations across images, the pre-trained frozen VFM fails to match the cross-view images. Additionally, the full-parameter supervised fine-tuning (SFT) approach cannot fully exploit the potential of VFMs, and may even harm its general feature representation ability \cite{harm_network}, especially when training data are limited. As shown in Fig. \ref{fig:introductionmelf}, the frozen VFM and SFT model tend to focus on background regions (e.g., tree textures) and struggle to maintain consistent attention on object regions across viewpoints. Therefore, this paper aims to employ the general visual features of the VFM for boosting CVGL performance and use the parameter-efficient transfer learning (PETL) \cite{PETL} to effectively adapt VFM to the downstream CVGL task.

To this end, we propose a parameter-efficient adaptation framework for CVGL, \textbf{B}ridge the \textbf{G}eometric \textbf{G}ap (\textbf{BGG}). BGG can enhance the robustness and consistency of features for cross-view image matching and preserve the generalization capabilities of the VFM, thereby significantly improving CVGL performance. Specifically, BGG adopts the pre-trained DINOv3 as the feature extraction backbone and designs two components for bridging the geometric gap between cross-view images, such as a multi-granularity feature enhancement adapter (MFEA) and a frequency-aware structural aggregation (FASA) module. MFEA utilizes multi-level dilated convolutions to effectively enlarge the receptive field and capture multi-scale spatial relations. MFEA can enhance the scale adaptability and robustness of features to image view changes, effectively bridging the geometric gap across two views with minimal training costs. Secondly, FASA extracts discriminative local structural information by modulating patch tokens in the frequency domain and incorporates an adaptive aggregation mechanism. Finally, BGG concatenates the enhanced local features with the global \texttt{[CLS]} token for improving the accuracy of cross-view matching and localization.

The main contributions of this paper are summarized as follows:

\begin{itemize}
    \item We propose a parameter-efficient adaptation framework for CVGL, \textbf{B}ridge the \textbf{G}eometric \textbf{G}ap (\textbf{BGG}). BGG can bridge the geometric gap between cross-view images and extract consistent features for CVGL, which can efficiently leverage the powerful representation capabilities and generalization of the VFM for boosting CVGL performance.

    \item We design a multi-granularity feature enhancement adapter (MFEA) to enlarge the receptive field and capture multi-scale spatial relations. It can enhance the robustness of feature representations to scale changes and geometric variations in cross-view images with minimal trainable parameters.

    \item We propose a frequency-aware structural aggregation (FASA) module to extract the discriminative local structural information by modulating patch tokens in the frequency domain. It highlights cross-view consistent and discriminative structural regions, improving the CVGL performance.

    \item Extensive experiments on benchmark datasets (University-1652 \cite{University} and SUES-200 \cite{SUES_200}) demonstrate that our BGG achieves state-of-the-art localization performance with fewer trainable parameters, demonstrating strong effectiveness and superior generalization in cross-view scenarios.
\end{itemize}

%\begin{itemize}
%    \item We propose a parameter-efficient adaptation framework tailored for the CVGL task to \textbf{B}ridge the \textbf{G}eometric \textbf{G}ap (\textbf{BGG}). BGG achieves efficient CVGL adaptation while retaining the powerful representation capabilities of the DINOv3 foundation model.

%    \item We design a multi-granularity feature enhancement adapter (MFEA). By leveraging multi-level dilated convolutions, it models multi-scale spatial relationships ranging from local textures to spatial contexts. This enhances the model's perception of cross-view scale variations and geometric distortions with minimal trainable parameters.

%    \item We propose a frequency-aware structural aggregation (FASA) module. This module introduces a frequency-aware mechanism combined with gated mixing and adaptive weighted aggregation strategies. It highlights cross-view consistent structural regions, improving the robustness and consistency of retrieval descriptors.

%    \item Extensive experiments on benchmark datasets (University-1652\cite{University} and SUES-200\cite{SUES_200}) demonstrate that our method achieves state-of-the-art localization performance with fewer trainable parameters, demonstrating strong effectiveness and superior generalization in cross-view scenarios.
%\end{itemize}

\section{Related Works}
This paper presents an efficient adaptation method for vision foundation models tailored to the Cross-view Geo-localization (CVGL) task, aiming to achieve effective CVGL adaptation while preserving the strong representation capability of the foundation model. To this end, we briefly review prior work on cross-view geo-localization and parameter-efficient transfer learning.

\subsection{Cross-View Geo-Localization}

Cross-View Geo-Localization (CVGL) aims to address image retrieval tasks under extreme viewpoint discrepancies (e.g., matching drone/ground images with satellite imagery).  It holds critical application value in GNSS-denied navigation and urban surveillance.  This field has evolved from ground-satellite matching to the more challenging drone-satellite matching. Early benchmark datasets, such as CVUSA \cite{CVUSA} and CVACT \cite{CVACT}, propelled the application of deep learning in this domain. Subsequently, VIGOR \cite{Vigor} introduced more realistic off-center alignment settings. With the rise of low-altitude remote sensing, University-1652 \cite{University}  and SUES-200 \cite{SUES_200} were proposed for drone viewpoints. These datasets bring more severe scale variations, multi-view sequences, and environmental disturbances. Consequently, this imposes an imperative demand for models with stronger geometric robustness. 

To bridge the cross-view geometric gap, research has transitioned from explicit geometric transformations to implicit feature alignment. Early methods used polar coordinate transforms \cite{SAFA}, perspective projection \cite{PCL}, or GAN-based generation \cite{Coming_down} to physically align viewpoints. However, explicit transformations may introduce artifacts. Thus, subsequent works moved to implicit alignment. Specifically, CVFT \cite{CVFT}  uses optimal transport to establish cross-view feature mappings. F3-Net \cite{F3_Net} and DAC \cite{DAC} align feature spaces via distribution metrics and pixel-level constraints, respectively. In addition, data-driven strategies, such as hard negative mining \cite{Sample4geo} and specific data augmentation \cite{LCM, GeoDTR, GeoDTR+} have been shown to effectively improve retrieval discriminability.

At the architectural level, CNNs are constrained by their local receptive fields. Consequently, recent works have gradually shifted towards Transformers \cite{L2LTR, FSRA, Transgeo, TransFG, SRLN} or hybrid architectures \cite{SAID, MCCG} to capture global context. For example, FSRA \cite{FSRA} and TransGeo \cite{Transgeo} use attention-based masks for region alignment. SRLN \cite{SRLN} adopts the Swin Transformer and better handles multi-scale variations. For descriptor construction, the research focus has shifted from generic aggregation operators (e.g., NetVLAD \cite{NetVLAD}, SAFA \cite{SAFA}, and GeM  \cite{GeM} ) to more fine-grained representations. Typical approaches such as region partitioning \cite{LPN}, keypoint-guided modeling \cite{RK_net}, and joint optimization of global and local features \cite{MEAN} are employed to enhance the perception of spatial layouts.

% However, despite the notable progress, existing methods still face two key bottlenecks. First, feature extraction networks suffer from limited generalization capability. Most approaches rely on complex task-specific architectures, leading to high training overhead and limited generalization capabilities.
% Currently, the visual foundation model DINOv3 \cite{Dinov3} possesses powerful general representation capabilities.  However, due to the inherent discrepancies in data distribution and geometric objectives between pre-training and the CVGL task, direct application or full fine-tuning often yields suboptimal results.  To bridge this gap,  we propose BGG, an efficient adaptation framework of vision foundation models for CVGL. BGG preserves the semantic capability of the foundation model while enabling targeted geometric adaptation with minimal parameter overhead. Second, there is the issue of structural information loss and noise interference. 
Although existing methods have achieved notable progress, they still have several limitations. Most current approaches rely on complex, specialized architectures, incurring substantial training and inference overheads while often exhibiting limited generalization in large-scale or highly dynamic scenarios. Currently, Vision Foundation Models (VFMs) like DINOv3 possess powerful general representation capabilities. However, due to severe geometric variations across views, direct application or full-parameter fine-tuning of these models often proves ineffective. To address this, we propose a parameter-efficient adaptation framework for the CVGL task, named Bridging the Geometric Gap (BGG). BGG aims to leverage the general visual features of the pre-trained DINOv3 for CVGL, while enhancing the adaptability of VFM to cross-view variations. Furthermore, existing transformer-based methods overly rely on the global \texttt{[CLS]} token, which tends to compress the local structural details essential for precise localization. Meanwhile, existing spatial domain aggregation struggles to effectively separate stable structures from high-frequency noise. Inspired by the success of frequency domain learning  \cite{MFAF, PIIS} in structural enhancement, we further propose the frequency-aware structural aggregation (FASA) module. FASA leverages frequency-domain modulation combined with gated mixing and adaptive weighted aggregation to extract robust structural information from patch tokens, thereby compensating for the limitations of global representations.

\subsection{Parameter-Efficient Transfer Learning}
The rise of Visual Foundation Models (VFMs) \cite{Dinov3, MAE, CLIP} has revolutionized the computer vision paradigm. Benefiting from large-scale self-supervised pre-training, models like DINOv3 \cite{Dinov3} possess powerful general semantic representations and out-of-distribution robustness. However, full fine-tuning of these massive models is not only computationally expensive but also prone to destroying the general knowledge structure acquired during pre-training. This issue is particularly evident for CVGL, where training data are often limited.

To address this challenge, PETL originates from NLP (e.g., Adapters \cite{PETL} and LoRA  \cite{Lora} ). Its core idea is to freeze the pre-trained backbone and adapt to downstream tasks by optimizing only a minimal number of additional parameters. Although methods like AdaptFormer \cite{Adaptformer} and Mv-Adapter \cite{Mv_adapter} have successfully transferred this paradigm to general visual tasks, their application in the CVGL domain remains in an exploratory stage \cite{CVGL_adapter}. A critical limitation is that most existing PETL modules are originally designed for language-oriented sequential modeling \cite{Elp_adapter, vmt_adapter} and are inherently spatially unaware. As a result, they lack the geometric awareness required to handle severe geometric distortions in CVGL, and they often ignore the multi-scale structural features that are crucial for matching.

To address these challenges, we propose an efficient adaptation framework for CVGL to bridge the geometric gap (BGG). Unlike generic PETL methods, BGG uses DINOv3 as the backbone and introduces a multi-granularity feature enhancement adapter (MFEA). MFEA discards the simple bottleneck structure. Instead, it utilizes multi-level dilated convolutions to expand the receptive field and models multi-scale spatial relationships ranging from local textures to spatial contexts. This enables BGG to enhance the foundation model's ability to handle drastic cross-view scale variations and geometric distortions with minimal parameter overhead.

\section{Methodology}
In this section, we first present the problem definition of Cross-View Geo-Localization (CVGL) (\ref{sec:Problem Formulation}). Subsequently, we provide an overview of our overall network architecture (\ref{sec:Overall Framework}). We then elaborate on two pivotal components: the multi-granularity feature enhancement adapter (MFEA), which performs efficient adaptation on the frozen DINOv3 backbone via multi-level dilated convolutions to handle multi-scale geometric variations (\ref{sec:Adapted DINOv3}); and the frequency-aware structural aggregation (FASA) module, which strengthens spatial structural information of features (\ref{sec:FASA}). Finally, we present the objective function utilized for model optimization (\ref{sec:Loss}).

\subsection{Problem Formulation}
\label{sec:Problem Formulation}
Cross-view geo-localization can be formulated as an image retrieval problem. Given two image sets captured from different viewpoints of the same location. let $\mathcal{R} = \{I_r^i\}_{i=1}^{M}$ denote the satellite reference database and $\mathcal{Q} = \{I_q^j\}_{j=1}^{N}$ represent the drone query set. Here, $I$ denotes an image, and $M$ and $N$ are the sizes of the reference database and the query set, respectively. Our objective is to learn a feature mapping function $\mathcal{F}(\cdot; \theta)$ that projects the visually discrepant $I_r$ and $I_q$ into a shared embedding space. Within this latent space, the similarity between descriptors implies their geographical proximity. Specifically, for a matched pair $(I_r, I_q)$, the cosine similarity is expected to be significantly higher than that of unmatched pairs.
In the inference phase, given a drone query image $I_q$, we compute its feature similarity to all satellite images in the database and retrieve the most similar one to determine its location:
\begin{equation}
L(I_q)=L(I_r^{\hat{k}}),  
\text{where } \hat{k}=\arg\max_{k}\operatorname{sim}\!\big(\mathcal{F}(I_q), \mathcal{F}(I_r^{k})\big),
\end{equation}
where $L(\cdot)$ denotes the location function and $\operatorname{sim}(\cdot)$ is the cosine similarity.

\begin{figure*}[htbp]  
    \centering
    \includegraphics[width=\textwidth]{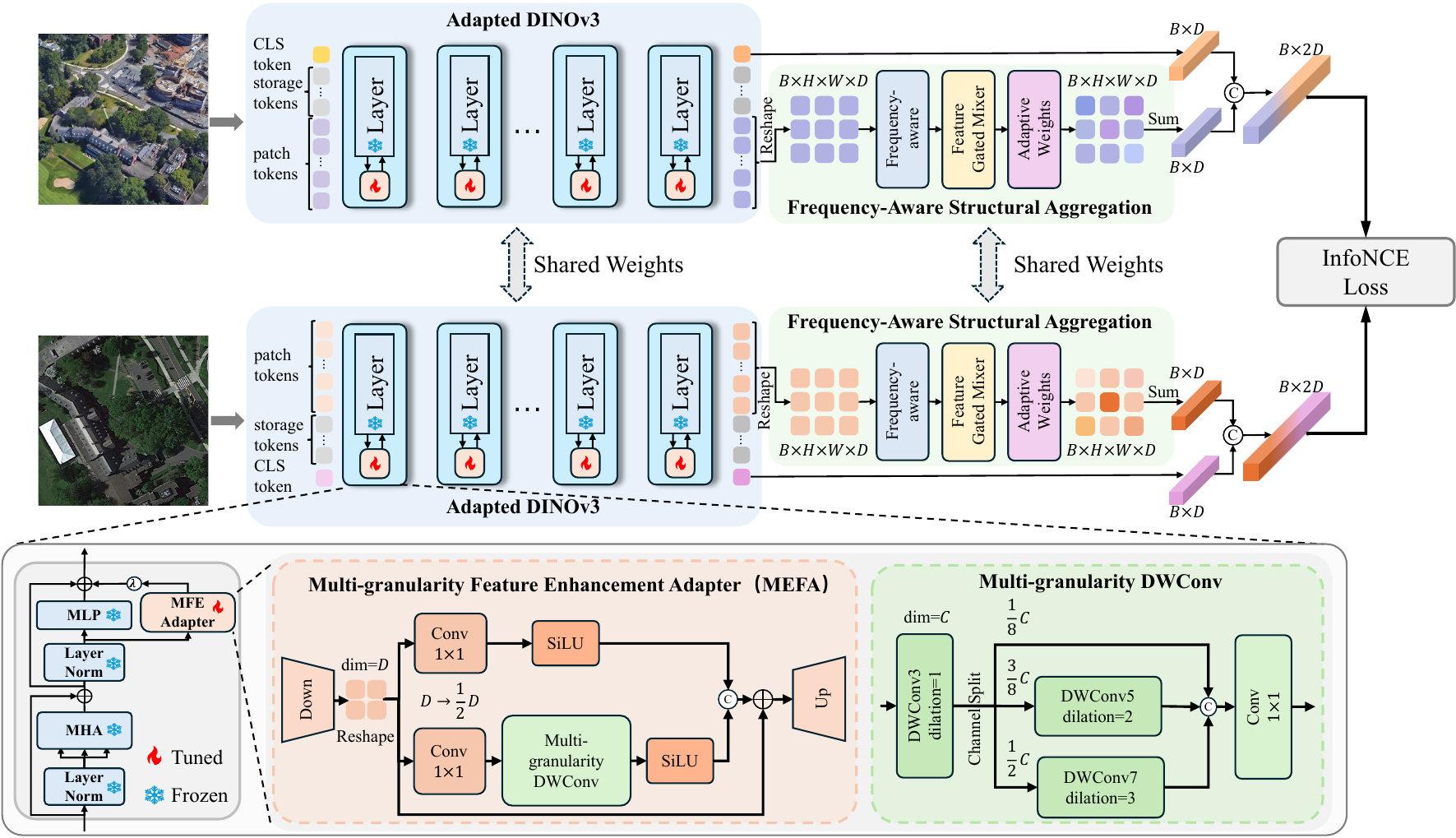}
    \caption{Overview of the proposed efficient adaptation framework of a vision foundation model for CVGL between drone and satellite imagery (BGG). BGG employs DINOv3 as the backbone and keeps the backbone parameters frozen.  We insert a multi-granularity feature enhancement adapter (MFEA, shown in the orange block) into the transformer blocks to perform efficient adaptation. MFEA employs multi-level dilated convolutions to model multi-scale spatial relations from local textures to broader spatial context. 
    After the backbone network extracts features, BGG applies a frequency-aware structural aggregation (FASA) module on patch tokens to extract more stable local structural information. Finally, we concatenate the \texttt{[CLS]} token with the aggregated local feature to form the image descriptor, and optimize the network using the InfoNCE loss.}
   \label{fig:pipeline}
\end{figure*}

\subsection{Overall Framework}
\label{sec:Overall Framework}
% Figure~\ref{fig:pipeline} illustrates our proposed BGG, an efficient adaptation framework for DINOv3\cite{Dinov3} foundation model tailored for CVGL between drone and satellite imagery. 
Figure~\ref{fig:pipeline} illustrates our proposed efficient adaptation framework for CVGL, termed BGG. The overall framework primarily consists of an adapted DINOv3 feature extraction network followed by a subsequent frequency-aware structural aggregation (FASA) module. Specifically, 
BGG utilizes DINOv3 as the backbone, where we freeze its parameters and inject a multi-granularity feature enhancement adapter (MFEA) in parallel within the transformer blocks. 
By exploiting multi-level dilated convolutions, MFEA models multi-scale spatial relationships ranging from local textures to broader spatial contexts, thereby endowing the frozen foundation model with scale adaptivity and cross-view robustness at a minimal parameter overhead. 
Furthermore, to compensate for the deficiency of the \texttt{[CLS]} token in representing spatial details, we propose the FASA module. 
FASA combines frequency-aware modulation with gated mixing and adaptive weighting to extract more stable local structural information. The overall pipeline can be formulated as:

\begin{align}
\mathbf{c} &= \operatorname{AdaptedDINOv3}(I)_{\mathrm{cls}} \in \mathbb{R}^{D}, \\
\mathbf{P} &= \operatorname{AdaptedDINOv3}(I)_{\mathrm{patch}} \in \mathbb{R}^{N \times D}, \\
\mathbf{p} &= \operatorname{FASA}(\mathbf{P}) \in \mathbb{R}^{D}, \\
\mathbf{z} &= \operatorname{Concat}(\mathbf{c}, \mathbf{p}) \in \mathbb{R}^{2D},
\end{align}
where $I$ denotes the input image, $\operatorname{AdaptedDINOv3}$ is DINOv3 adapted with MFEA, 
$\mathbf{c}$ and $\mathbf{P}$ are the \texttt{[CLS]} token and patch tokens, respectively, 
$D$ is the feature dimension and $\operatorname{Concat}$ denotes concatenation.

In this framework, the drone and satellite branches share the weights of the adapted backbone as well as the FASA module. The whole deep model is optimized end-to-end using the InfoNCE loss.

\subsection{Adapted DINOv3}
\label{sec:Adapted DINOv3}
Recently, Visual Foundation Models (VFMs) \cite{Adaptformer, Conv_adapter} have exhibited powerful general semantic representation capabilities. For instance, DINOv3 \cite{Dinov3} learns universal and high-quality global and dense feature representations through self-supervised pretraining, and achieves remarkable performance across various vision tasks. However, due to the significant discrepancies in training data and objectives between pre-training and CVGL tasks, direct application often fails to effectively cope with the drastic scale variations and geometric deformations inherent in CVGL. To bridge this gap, we design the multi-granularity feature enhancement adapter (MFEA). It aims to enhance DINOv3's perception of cross-view geometric transformations by injecting multi-granularity geometric spatial priors, without compromising the pre-trained knowledge.

We employ DINOv3 (ViT-B/16) as the backbone network, which is primarily distilled from a ViT-7B model trained via self-supervision on the LVD-1689M dataset. 
Given an input image $I \in \mathbb{R}^{3 \times H \times W}$, patch embedding produces an input token sequence $X_0 \in \mathbb{R}^{(L+1) \times D}$, where $L$ is the number of patch tokens and $D$ denotes the feature dimension. To enable DINOv3 to perceive the multi-granularity spatial relationships required for CVGL, we insert the lightweight MFEA module in parallel into each Transformer block. The computation of the $l$-th block is formulated as:

\begin{align}
X'_l &= X_{l-1} + \operatorname{MHA}(\operatorname{LN}(X_{l-1})), \\
X_l  &= X'_l + \operatorname{FFN}(\operatorname{LN}(X'_l)) 
      + \lambda \cdot \operatorname{MFEA}_l(\operatorname{LN}(X'_l)),
\end{align}
where $\operatorname{MHA}$ and $\operatorname{FFN}$ denote the multi-head self-attention and feed-forward network with frozen parameters, respectively, 
$\operatorname{MFEA}_l$ is the trainable lightweight MFEA module at layer $l$, 
$\operatorname{LN}$ denotes layer normalization, and $\lambda$ is a scaling factor that modulates the adapter output.

In parameter-efficient transfer learning, standard adapters typically adopt a ``bottleneck'' architecture, comprising a down-projection layer (fully connected layer), a non-linear activation function, and an up-projection layer. To better adapt DINOv3 to the scale variations and geometric deformations inherent in CVGL scenarios, we introduce a multi-granularity feature enhancement module while retaining the bottleneck structure (i.e., the down-/up-projection layers). Specifically, the input sequence $X_l \in \mathbb{R}^{(L+1) \times D}$ is first projected onto a lower dimension $D'$ via a fully connected layer to reduce parameter overhead. We then reshape the patch tokens into a 2D feature map $\hat{X} \in \mathbb{R}^{B \times D' \times H \times W}$ to recover spatial structures.

As illustrated in Fig. \ref{fig:pipeline}, MFEA employs a dual-branch architecture. The first branch utilizes a $1 \times 1$ convolution followed by a $\operatorname{SiLU}$ activation to refine local texture information while maintaining a lightweight design (reducing the feature dimension to $D'/2$).  The second branch similarly maps the feature dimension to $D'/2$ via a $1 \times 1$ convolution, followed by multi-granularity depthwise separable convolutions to capture multi-scale spatial relationships. Specifically, we first employ a depthwise separable convolution with a kernel size of $3 \times 3$ and dilation rate $d=1$ to extract base features. The features are then split along the channel dimension into three groups $\{V_1, V_2, V_3\}$ with a ratio of $[1:3:4]$. Subsequently, we apply depthwise separable convolutions with different dilation rates to capture multi-granularity receptive fields. Finally, we concatenate the resulting features along the channel dimension and use a $1\times1$ convolution to fuse inter-granularity relations. This process can be formulated as:

\begin{align}
&\hat{X}_{1} = \operatorname{Conv}_{1\times 1}(\hat{X}), \\
&X_{\text{b1}} = \operatorname{SiLU}(\hat{X}_{1}), \\
&\hat{X}_{2} = \operatorname{DWConv}_{3\times3,\,d=1}(\operatorname{Conv}_{1\times1}(\hat{X})), \\
&Y_{1} = \operatorname{Identity}(V_{1}), \\
&Y_{2} = \operatorname{DWConv}_{5\times5,\,d=2}(V_{2}), \\
&Y_{3} = \operatorname{DWConv}_{7\times7,\,d=3}(V_{3}), \\
&X_{\text{b2}} = \operatorname{SiLU}(\operatorname{Conv}_{1\times1}(\operatorname{Concat}(Y_1, Y_2, Y_3))),
\end{align}
where $\operatorname{DWConv}_{l\times l}$ denotes a depthwise separable convolution with kernel size $l\times l$, $d$ is the dilation rate, $\operatorname{Identity}$ denotes the identity mapping, $\text{b1}$ and $\text{b2}$ represent the first branch and the second branch, respectively.

Subsequently, the outputs of the two branches are concatenated along the channel dimension and fused with the input features via a residual connection to obtain a feature map enriched with multi-scale spatial information. We then flatten the feature map and use an up-projection layer to restore the original feature dimension $D$. Finally, the result is concatenated with the \texttt{[CLS]} token to form the adapter output sequence $\hat{X}'_l \in \mathbb{R}^{(L+1)\times D}$, which is formulated as follows:
\begin{align}
X_{\text{flt}} &= \operatorname{Flatten}(\operatorname{Concat}(X_{\text{b1}}, X_{\text{b2}})), \\
\hat{X}'_l &= \operatorname{Linear}_{\text{up}}(X_{\text{flt}}),
\end{align}
where $\operatorname{Flatten}$ denotes the feature-map flattening operation and $\operatorname{Linear}_{\text{up}}$ represents the up-projection layer.

% Overall, MFEA effectively models multi-scale spatial relationships ranging from local textures to spatial contexts while incurring minimal parameter overhead. This significantly enhances the model's adaptability to cross-view scale variations and geometric distortions.

\begin{figure}[tbp]  
    \centering
    \includegraphics[width=\linewidth]{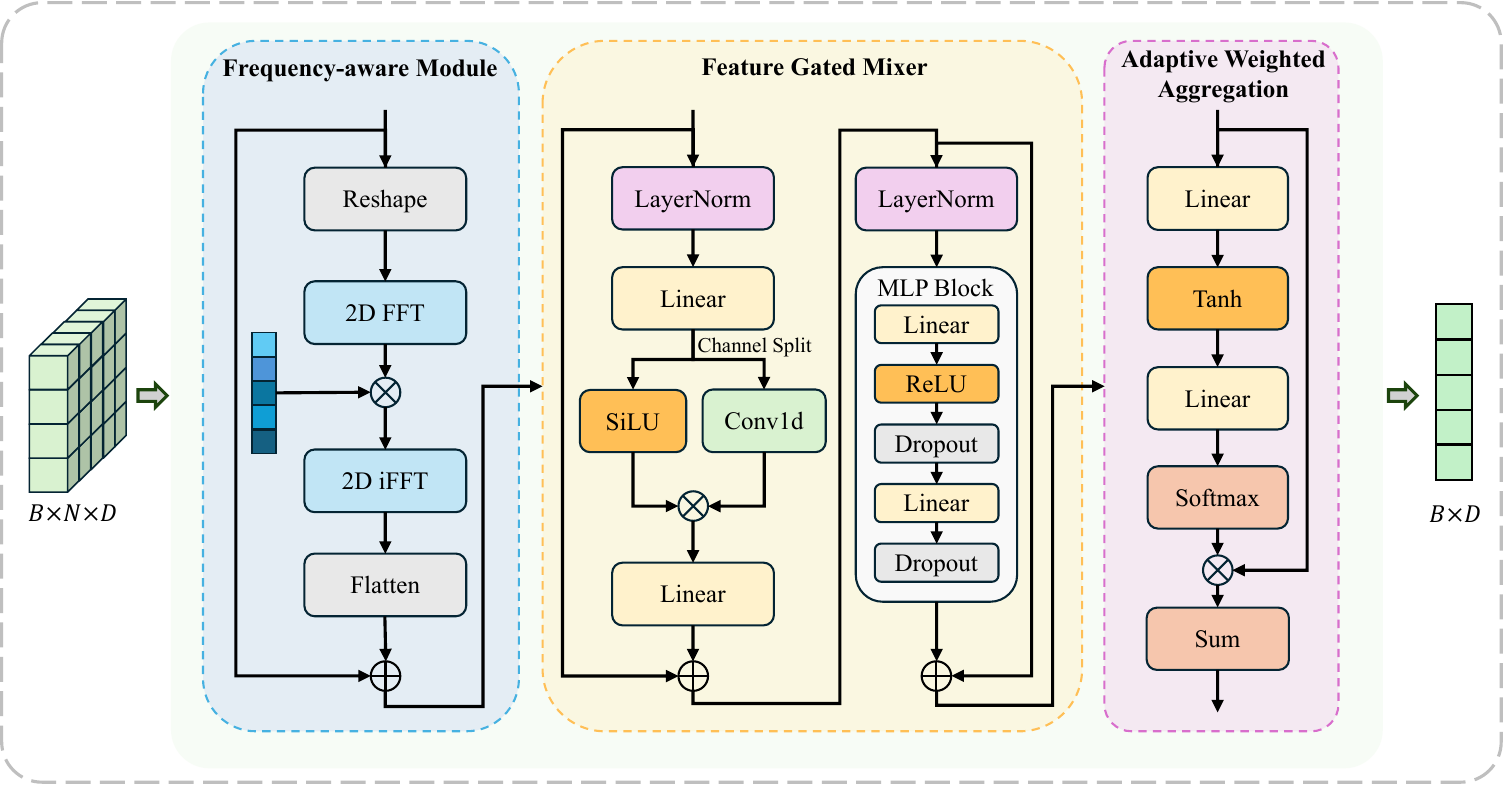}
    \caption{Overview of the proposed frequency-aware structural aggregation (FASA) module. FASA combines frequency-aware modulation with gated mixing and adaptive weighting to extract more stable local structural information, and finally outputs an aggregated feature vector. }
   \label{fig:FASA}
\end{figure}

\subsection{Frequency-Aware Structural Aggregation}
\label{sec:FASA}
In retrieval tasks, the highly compressed \texttt{[CLS]} token is commonly utilized as a global descriptor. However, it often weakens crucial spatial layout information in localization scenarios. To extract more robust spatial structural features, we propose a frequency-aware structural aggregation module (FASA). FASA can enhance stable local structural information by integrating frequency-aware modulation with feature gated mixing and adaptive weighted aggregation.

Cross-view discrepancies are often more pronounced in high-frequency details (e.g., textures and shadows), whereas structural information, such as roads and building edges are relatively more stable. 
Consequently, FASA applies frequency-aware modulation to the patch tokens. This process adaptively enhances structural frequency responses that exhibit cross-view consistency while suppressing view-sensitive interfering frequency components. As illustrated in Fig. \ref{fig:FASA}, following feature extraction by the backbone, the patch token sequence $X_p$ is reshaped into a 2D feature map and transformed into the frequency domain via the 2D Fast Fourier Transform (FFT).  We introduce a learnable complex weight matrix $W_{freq} \in \mathbb{R}^{H \times W}$ to modulate the spectrum, and then transform it back to the spatial domain with an inverse transform. Finally, the feature map is flattened and added to the input sequence through a residual connection.  The process is formulated as:

\begin{align}
    X_{freq} &= \operatorname{iFFT}(\operatorname{FFT}(X_p) \odot W_{freq}), \\
    X_{FM} &= \operatorname{Flatten}(X_{freq}) + X_{p},
\end{align}
where $\operatorname{FFT}$ and $\operatorname{iFFT}$ denote the fast Fourier transform and its inverse, respectively, and $\odot$ denotes element-wise multiplication.

To facilitate feature interaction across channel and spatial dimensions, we design a feature gated mixer. 
The features are first normalized via $\operatorname{LayerNorm}$ and projected into a higher-dimensional space. The projected features are then split along the channel dimension into a content component $X_{\text{c}}$ and a gating component $X_{\text{g}}$. The content component models local dependencies via a 1D convolution, while the gating component produces activation weights through $\operatorname{SiLU}$.  After their interaction, an $\operatorname{MLP}$ is applied for non-linear feature fusion. The process is formulated as:

\begin{align}
    &X_{c}, X_{g} = \operatorname{Chunk}(\operatorname{Linear}(X_{FM})), \\
    &X_{mid} = \operatorname{Linear}(\operatorname{Conv1d}(X_{c}) \odot \operatorname{SiLU}(X_{g})) + X_{FM}, \\
    &X_{mix} = \operatorname{MLP}(\operatorname{LayerNorm}(X_{mid})) + X_{mid}.
\end{align}

Finally, to better focus on key regions exhibiting cross-view consistency, rather than employing simple average pooling, we utilize a lightweight network to predict spatial importance weights $W_{att} \in \mathbb{R}^{L \times 1}$ for each patch token and perform weighted aggregation on $X_{mix}$. This process is formulated as follows:

\begin{align}
    &W_{att} = \operatorname{Softmax}(\operatorname{Linear}(\tanh(\operatorname{Linear}(X_{mix})))), \\
    &X_{FASA} = \sum_{i=1}^{L} W^{i}_{att} \cdot X_{mix}^{(i)},
\end{align}
where $\tanh$ is the activation function. 

We concatenate $X_{FASA}$ with the \texttt{[CLS]} token to form an image descriptor that captures both global semantics and stable spatial structural information.

\subsection{Loss Function}
\label{sec:Loss}

To preserve cross-view consistency and discriminability, we adopt the InfoNCE loss as the optimization objective. By contrasting embeddings from the same location (positive pairs) against embeddings from different locations (negative pairs),  InfoNCE loss pulls matched drone-satellite pairs closer in the embedding space while pushing mismatched pairs apart. Assuming a training batch contains $B$ matched image pairs, the loss is formulated as follows:

\begin{equation}
    \mathcal{L}_{q\to r} = -\frac{1}{B}\sum_{i=1}^{B}\log\frac{\exp\!\left(\operatorname{sim}(q_{i},r_{+})/\tau \right)}{\sum_{j=1}^{B}\exp\!\left(\operatorname{sim}(q_{i},r_{j})/\tau\right)},
\end{equation}
where $q_i$ denotes the encoded feature of the $i$-th query image, $r_i^{+}$ is the feature of its matched reference image, $\operatorname{sim}(\cdot)$ denotes cosine similarity, and $\tau$ is a learnable temperature parameter that controls the sharpness of the distribution. To strengthen bidirectional retrieval, we use a symmetric InfoNCE formulation and compute losses for both drone-satellite and satellite-drone directions. The final objective is:

\begin{equation}
    \mathcal{L} = \frac{1}{2}(\mathcal{L}_{u\to s} + \mathcal{L}_{s\to u}),
\end{equation}
where $u$ and $s$ denote drone and satellite images, respectively. 
By minimizing this objective, the model learns discriminative, viewpoint-invariant features for accurate cross-view geo-localization.

\begin{table*}[ht]
  \centering
  \caption{Performance comparison with state-of-the-art methods on the University-1652 dataset. The best and second-best results are highlighted in \textcolor{red}{\textbf{red}} and \textcolor{blue}{\textbf{blue}}, respectively. ``*'' indicates methods employing Content-Viewpoint Disentanglement~\cite{CVD} during training.}
  \resizebox{0.9\linewidth}{!}{
    % \begin{tabular}{lcccccc}
    \begin{tabular}{lcc
                >{\hspace{10pt}}c<{\hspace{8pt}}
                >{\hspace{8pt}}c<{\hspace{10pt}}
                >{\hspace{8pt}}c<{\hspace{8pt}}
                >{\hspace{8pt}}c<{\hspace{10pt}}}
    \toprule
    \multicolumn{1}{c}{\multirow{2}[4]{*}{Model}} & \multirow{2}[4]{*}{Parameters (M)} & \multirow{2}[4]{*}{GFLOPs} & \multicolumn{2}{c}{Drone$\rightarrow$Satellite} & \multicolumn{2}{c}{Satellite$\rightarrow$Drone} \\
    % \multirow{2}[4]{*}{Model} & \multirow{2}[4]{*}{Parameters (M)} & \multirow{2}[4]{*}{GFLOPs} & \multicolumn{2}{c}{Drone$\rightarrow$Satellite} & \multicolumn{2}{c}{Satellite$\rightarrow$Drone} \\
\cmidrule{4-7}          &       &       & R@1   & AP    & R@1   & AP \\
    \midrule
    MuSe-Net \cite{university_weather_MuSeNet} & 50.47  & -     & 74.48  & 77.83  & 88.02  & 75.10  \\
    LPN \cite{LPN}   & 62.39  & \color{blue}{36.78} & 75.93  & 79.14  & 86.45  & 74.49  \\
    F3-Net \cite{F3_Net} & -     & -     & 78.64  & 81.60  & -     & - \\
    TransFG \cite{TransFG} &$>$86.00 & -     & 84.01  & 86.31  & 90.16  & 84.61  \\
    IFSs \cite{IFSs}  & -     & -     & 86.06  & 88.08  & 91.44  & 85.73  \\
    MCCG \cite{MCCG}  & 56.65  & 51.04  & 89.40  & 91.07  & 95.01  & 89.93  \\
    SDPL \cite{SDPL}  & 42.56  & 69.71  & 90.16  & 91.64  & 93.58  & 89.45  \\
    MFJR \cite{MFJR}  & $>$88.00 & -     & 91.87  & 93.15  & 95.29  & 91.51  \\
    CCR \cite{CCR}   & 156.57  & 160.61  & 92.54  & 93.78  & 95.15  & 91.80  \\
    Sample4Geo cite{Sample4geo} & 87.57  & 90.24  & 92.65  & 93.81  & 95.14  & 91.39  \\
    SRLN \cite{SRLN}  & 193.03  & -     & 92.70  & 93.77  & 95.14  & 91.97  \\
    DAC \cite{DAC}   & 96.50  & 90.24  & \color{blue}{94.67} & \color{blue}{95.50} & \color{blue}{96.43} & \color{blue}{93.79} \\
    Game4Loc \cite{Game4loc} & -     & -     & 91.32  & 92.56  & 94.43  & 90.83  \\
    Game4Loc* \cite{CVD} & -     & -     & 92.94  & 93.95  & 94.86  & 91.92  \\
    MEAN \cite{MEAN}  & \color{blue}{36.50} & \color{red}{26.18} & 93.55  & 94.53  & 96.01  & 92.08  \\
    \midrule
    Ours  & \color{red}{10.67} & 86.63  & \color{red}{96.24} & \color{red}{96.81} & \color{red}{97.57} & \color{red}{95.44} \\
    \bottomrule
    \end{tabular}%
    }
  \label{tab:university}%
\end{table*}%

\section{Experiments and results}
In this section, we evaluate the effectiveness and advantages of the proposed parameter-efficient adaptation framework, BGG, for cross-view geo-localization. We first describe the datasets and experimental settings. Subsequently, we conduct a comparative analysis of the proposed method against current representative approaches. Finally, we conduct qualitative visualizations and ablation studies to analyze the contribution of each component.

\subsection{Datasets and Evaluation Protocols}

% This work focuses on cross-view localization between drone and satellite imagery. 
To comprehensively evaluate the effectiveness and robustness of our method under diverse challenges, we conduct experiments on three mainstream datasets: University-1652  \cite{University}, SUES-200 \cite{SUES_200}, and Multi-weather University-1652  \cite{university_weather_MuSeNet}. These benchmarks provide complementary evaluation scenarios, covering multi-view settings, multi-scale discrepancies, and extreme environmental conditions, respectively. Together, they offer a thorough testbed for drone-satellite cross-view geo-localization.

\textbf{University-1652} is the first multi-view benchmark dataset for drone-based visual geo-localization, encompassing images from drone, satellite, and ground views. It covers 1,652 distinct locations across 72 universities worldwide. The training set contains 701 locations from 33 universities, while the test set contains 951 locations from 39 universities. Each location provides 54 drone images captured at different altitudes and viewing angles, one satellite image, and multiple ground view images. 
% This dataset supports two tasks: drone target localization and autonomous navigation.

\textbf{SUES-200} is a recently released dataset explicitly designed to evaluate model robustness to drone altitude changes and drastic scale discrepancies. Unlike previous benchmarks, SUES-200 includes drone view images captured at four altitudes: 150m, 200m, 250m, and 300m. The dataset collects 200 locations, with 120 for training and 80 for testing. Each location provides one satellite view image and 50 drone view images. Beyond campus buildings, the scenes extend to parks, lakes, and urban infrastructure. In our experiments, by analyzing localization performance across different altitudes, the dataset enables assessment of a model's adaptability to scale variations and viewpoint differences.

\textbf{Multi-weather University-1652} is an extended version of the University-1652 dataset. It simulates diverse adverse environmental conditions, such as fog, rain, and snow, through 10 carefully designed data augmentation settings. This dataset is used to evaluate model robustness under extreme environmental conditions.

\textbf{Evaluation Protocol.} Following prior works \cite{MEAN, DAC}, we evaluate performance using Recall@K (R@K) and average precision (AP). R@K is defined as the percentage of queries whose correct match appears in the Top-$K$ retrieved results. A higher R@K indicates superior localization performance. AP measures the trade-off between precision and recall and is computed as the area under the precision-recall curve. A higher AP implies better overall performance.

\subsection{Implementation Details}

We employ DINOv3 (ViT-B/16) as the backbone network, which is primarily distilled from a ViT-7B model pre-trained via self-supervision on the LVD-1689M dataset. The backbone is kept frozen throughout training, and we only optimize the trainable parameters of the MFEA and FASA modules. The scaling factor $\lambda$ in MFEA is set to $0.9$. The output feature dimension of the adapted backbone is 768, resulting in a final feature dimension of 1536, obtained by concatenating the \texttt{[CLS]} token and the aggregated patch features along the channel dimension.

During the data preprocessing phase, all input images are resized to $3 \times 384 \times 384$. To enhance generalization, we apply data augmentations during training, including random cropping, color-space adjustment, and blurring. We use AdamW as the optimizer with an initial learning rate of $0.001$, and set the batch size to 64 (each batch contains 32 drone images and 32 satellite images). The temperature parameter $\tau$ in the InfoNCE loss is learnable and is initialized to $0.07$. We train the model for 5 epochs. At inference time, cosine similarity between feature vectors is used as the metric for retrieval and localization.

Our method is implemented using the PyTorch deep learning framework. All experiments are conducted on  NVIDIA GeForce RTX 3090 GPUs with 24GB of memory.

\begin{table*}[ht]
  \centering
  \caption{Performance comparison with state-of-the-art methods on the SUES-200 dataset under the ``Drone $\rightarrow$ Satellite'' setting. The best and second-best results are highlighted in \textcolor{red}{\textbf{red}} and \textcolor{blue}{\textbf{blue}}, respectively. ``*'' indicates methods employing Content-Viewpoint Disentanglement \cite{CVD} during training.}
  \resizebox{\linewidth}{!}{
    \begin{tabular}{lcccccccccc}
    \toprule
    \multicolumn{1}{c}{\multirow{3}[6]{*}{Model}} & \multirow{3}[6]{*}{Parameters(M)} & \multirow{3}[6]{*}{GFLOPs} & \multicolumn{8}{c}{Drone$\rightarrow$Satellite} \\
    % \multirow{3}[6]{*}{Model} & \multirow{3}[6]{*}{Parameters(M)} & \multirow{3}[6]{*}{GFLOPs} & \multicolumn{8}{c}{Drone$\rightarrow$Satellite} \\
\cmidrule{4-11}          &       &       & \multicolumn{2}{c}{150m} & \multicolumn{2}{c}{200m} & \multicolumn{2}{c}{250m} & \multicolumn{2}{c}{300m} \\
\cmidrule{4-11}          &       &       & R@1   & AP    & R@1   & AP    & R@1   & AP    & R@1   & AP \\
    \midrule
    LPN \cite{LPN}   & 62.39  & \color{blue}{36.78} & 61.58  & 67.23  & 70.85  & 75.96  & 80.38  & 83.80  & 81.47  & 84.53  \\
    IFSs \cite{IFSs}  & -     & -     & 77.57  & 81.30  & 89.50  & 91.40  & 92.58  & 94.21  & 97.40  & 97.92  \\
    MCCG \cite{MCCG}  & 56.65  & 51.04  & 82.22  & 85.47  & 89.38  & 91.41  & 93.82  & 95.04  & 95.07  & 96.20  \\
    SDPL \cite{SDPL}  & 42.56  & 69.71  & 82.95  & 85.82  & 92.73  & 94.07  & 96.05  & 96.69  & 97.83  & 98.05  \\
    CCR \cite{CCR}   & 156.57  & 160.61  & 87.08  & 89.55  & 93.57  & 94.90  & 95.42  & 96.28  & 96.82  & 97.39  \\
    MFJR \cite{MFJR}  & $>$88.00 & -     & 88.95  & 91.05  & 93.60  & 94.72  & 95.42  & 96.28  & 97.45  & 97.84  \\
    SRLN \cite{SRLN}  & 193.03  & -     & 89.90  & 91.90  & 94.32  & 95.65  & 95.92  & 96.79  & 96.37  & 97.21  \\
    Sample4Geo \cite{Sample4geo} & 87.57  & 90.24  & 92.60  & 94.00  & 97.38  & 97.81  & 98.28  & 98.64  & 99.18  & 99.36  \\
    DAC \cite{DAC}   & 96.50  & 90.24  & \color{blue}{96.80} & \color{blue}{97.54} & 97.48  & 97.97  & 98.20  & 98.62  & 97.58  & 98.14  \\
    Game4Loc \cite{Game4loc} & -     & -     & 94.62  & 95.59  & 96.55  & 97.27  & 97.55  & 98.16  & 97.67  & 98.24  \\
    Game4Loc* \cite{DAC} & -     & -     & 95.80  & 96.70  & 97.10  & 97.78  & 97.60  & 98.14  & 98.65  & 98.98  \\
    MEAN \cite{MEAN}  & \color{blue}{36.50} & \color{red}{26.18} & 95.50  & 96.46  & \color{blue}{98.38} & \color{blue}{98.72 } & \color{blue}{98.95} & \color{blue}{99.17} & \color{red}{99.52} & \color{red}{99.63} \\
    \midrule
    Ours  & \color{red}{10.67} & 86.63  & \color{red}{99.30} & \color{red}{99.46} & \color{red}{99.45} & \color{red}{99.55} & \color{red}{99.53} & \color{red}{99.63} & \color{blue}{99.25} & \color{blue}{99.38} \\
    \bottomrule
    \end{tabular}%
    }
  \label{tab:SUE-D2S}%
\end{table*}%

\begin{table*}[ht]
  \centering
  \caption{Performance comparison with state-of-the-art methods on the SUES-200 dataset under the ``Satellite $\rightarrow$ Drone'' setting. The best and second-best results are highlighted in \textcolor{red}{\textbf{red}} and \textcolor{blue}{\textbf{blue}}, respectively. ``*'' indicates methods employing Content-Viewpoint Disentanglement \cite{CVD} during training.}
  \resizebox{\linewidth}{!}{
    \begin{tabular}{lcccccccccc}
    \toprule
    \multicolumn{1}{c}{\multirow{3}[6]{*}{Model}} & \multirow{3}[6]{*}{Parameters(M)} & \multirow{3}[6]{*}{GFLOPs} & \multicolumn{8}{c}{Satellite$\rightarrow$Drone} \\
    % \multirow{3}[6]{*}{Model} & \multirow{3}[6]{*}{Parameters(M)} & \multirow{3}[6]{*}{GFLOPs} & \multicolumn{8}{c}{Satellite$\rightarrow$Drone} \\
\cmidrule{4-11}          &       &       & \multicolumn{2}{c}{150m} & \multicolumn{2}{c}{200m} & \multicolumn{2}{c}{250m} & \multicolumn{2}{c}{300m} \\
\cmidrule{4-11}          &       &       & R@1   & AP    & R@1   & AP    & R@1   & AP    & R@1   & AP \\
    \midrule
    LPN \cite{LPN}    & 62.39  & \color{blue}{36.78} & 83.75  & 83.75  & 83.75  & 83.75  & 83.75  & 83.75  & 83.75  & 83.75  \\
    CCR \cite{CCR}    & 156.57  & 160.61  & 92.50  & 78.54  & 97.50  & 95.22  & 97.50  & 97.10  & 97.50  & 97.49  \\
    IFSs \cite{IFSs}  & -     & -     & 93.75  & 89.49  & 97.50  & 90.52  & 97.50  & 96.03  & \color{red}{100.00} & 97.66  \\
    MCCG \cite{MCCG}  & 56.65  & 51.04  & 93.75  & 89.72  & 93.75  & 92.21  & 96.25  & 96.14  & \color{blue}{98.75} & 96.64  \\
    SDPL \cite{SDPL}  & 42.56  & 69.71  & 93.75  & 83.75  & 96.25  & 92.42  & 97.50  & 95.65  & 96.25  & 96.17  \\
    SRLN \cite{SRLN}  & 193.03  & -     & 93.75  & 93.01  & 97.50  & 95.08  & 97.50  & 96.52  & 97.50  & 96.71  \\
    MFJR \cite{MFJR}  & $>$88.00 & -     & 95.00  & 89.31  & 96.25  & 94.72  & 94.69  & 96.92  & \color{blue}{98.75} & 97.14  \\
    Sample4Geo \cite{Sample4geo} & 87.57  & 90.24  & \color{blue}{97.50} & 93.63  & \color{blue}{98.75} & 96.70  & \color{blue}{98.75} & \color{blue}{98.28} & \color{blue}{98.75} & 98.05  \\
    DAC \cite{DAC}    & 96.50  & 90.24  & \color{blue}{97.50} & 94.06  & \color{blue}{98.75} & 96.66  & \color{blue}{98.75} & 98.09  & \color{blue}{98.75} & 97.87  \\
    Game4Loc \cite{Game4loc} & -     & -     & 93.75 & 93.06  & 96.25 & 94.50  & 96.25 & 94.92  & 95.00 & 95.36  \\
    Game4Loc* \cite{CVD} & -     & -     & 96.25 & 93.37  & 97.50 & 95.03  & 96.25 & 95.95  & 97.50 & 96.28  \\
    MEAN \cite{MEAN}  & \color{blue}{36.50} & \color{red}{26.18} & \color{blue}{97.50} & \color{blue}{94.75} & \color{red}{100.00} & \color{blue}{97.09} & \color{red}{100.00} & \color{blue}{98.28} & \color{red}{100.00} & \color{red}{99.21} \\
    \midrule
    Ours  & \color{red}{10.67} & 86.63  & \color{red}{98.75} & \color{red}{98.22} & \color{blue}{98.75} & \color{red}{98.24} & \color{blue}{98.75} & \color{red}{98.73} & \color{blue}{98.75} & \color{blue}{98.68} \\
    \bottomrule
    \end{tabular}%
    }
  \label{tab:SUE-S2D}%
\end{table*}%

% Table generated by Excel2LaTeX from sheet 'University-weather'
\begin{table*}[ht]
  \centering
  \caption{Performance comparison with state-of-the-art methods on the Multi-weather University-1652 dataset. The best and second-best results are highlighted in \textcolor{red}{\textbf{red}} and \textcolor{blue}{\textbf{blue}}, respectively.}
  \resizebox{\linewidth}{!}{
    \begin{tabular}{lcccccccccc}
    \toprule
    \multicolumn{1}{c}{\multirow{2}[2]{*}{Model}} & Normal & Fog   & Rain  & Snow  & Fog+Rain & Fog+Snow & Rain+Snow & Dark  & Over-exposure & Wind \\
          & R@1/AP & R@1/AP & R@1/AP & R@1/AP & R@1/AP & R@1/AP & R@1/AP & R@1/AP & R@1/AP & R@1/AP \\
    \midrule
          & \multicolumn{10}{c}{Drone$\rightarrow$Satellite} \\
    \midrule
    LPN \cite{LPN}   & 74.33/77.60 & 69.31/72.95 & 67.96/71.72 & 64.90/68.85 & 64.51/68.52 & 54.16/58.73 & 65.38/69.29 & 53.68/58.10 & 60.90/65.27 & 66.46/70.35 \\
    MuSeNet \cite{university_weather_MuSeNet} & 74.48/77.83 & 69.47/73.24 & 70.55/74.14 & 65.72/69.70 & 65.59/69.64 & 54.69/59.24 & 65.64/70.54 & 53.85/58.49 & 61.65/65.51 & 69.45/73.22 \\
    Sample4Geo \cite{Sample4geo} & 90.55/92.18 & 89.72/91.48 & 85.89/88.11 & 86.64/88.18 & 85.88/88.16 & 84.64/87.11 & 85.98/88.16 & 87.90/89.87 & 76.72/80.18 & 83.39/89.51 \\
    DAC \cite{DAC}   & \color{blue}{92.81}/\color{blue}{94.06} & \color{blue}{92.55}/\color{blue}{93.84} & \color{blue}{90.06}/\color{blue}{91.70} & \color{blue}{90.04}/\color{blue}{91.69} & \color{blue}{89.80}/\color{blue}{91.45} & \color{blue}{89.29}/\color{blue}{91.02} & \color{blue}{89.91}/\color{blue}{91.52} & \color{blue}{91.00}/\color{blue}{92.54} & \color{blue}{82.41}/\color{blue}{85.18} & \color{blue}{90.98}/\color{blue}{92.48} \\
    MEAN \cite{MEAN}  & 90.81/92.32 & 90.97/92.52 & 88.19/90.05 & 85.69/90.49 & 86.75/88.84 & 86.00/88.22 & 87.21/89.21 & 87.90/89.87 & 80.54/83.53 & 89.27/91.01 \\
    Ours  & \color{red}{96.49}/\color{red}{97.14} & \color{red}{95.81}/\color{red}{96.58} & \color{red}{93.78}/\color{red}{94.85} & \color{red}{94.14}/\color{red}{95.16} & \color{red}{92.64}/\color{red}{93.92} & \color{red}{92.63}/\color{red}{93.91} & \color{red}{92.52}/\color{red}{93.80} & \color{red}{96.04}/\color{red}{96.78} & \color{red}{88.33}/\color{red}{90.28} & \color{red}{95.32}/\color{red}{96.17} \\
    \midrule
          & \multicolumn{10}{c}{Satellite$\rightarrow$Drone} \\
    \midrule
    LPN \cite{LPN}   & 87.02/75.19 & 86.16/71.34 & 83.88/69.49 & 82.88/65.39 & 84.59/66.28 & 79.60/55.19 & 84.17/66.26 & 82.88/52.05 & 81.03/62.24 & 84.14/67.35 \\
    MuSeNet \cite{university_weather_MuSeNet} & 88.02/75.10 & 87.87/69.85 & 87.73/71.12 & 83.74/66.52 & 85.02/67.78 & 80.88/54.26 & 84.88/67.75 & 80.74/53.01 & 81.60/62.09 & 86.31/70.03 \\
    Sample4Geo \cite{Sample4geo} & 95.86/89.86 & 95.72/88.95 & 94.44/85.71 & 95.01/86.73 & 93.44/85.27 & 93.72/84.78 & 93.15/85.50 & 96.01/87.06 & 89.87/74.52 & 95.29/87.06 \\
    DAC \cite{DAC}   & \color{red}{97.43}/\color{blue}{92.90} & \color{red}{97.00}/\color{blue}{92.44} & 
    \color{blue}{95.72}/\color{blue}{90.00} & \color{blue}{96.01}/\color{blue}{90.48} & \color{blue}{95.29}/\color{blue}{89.52} & \color{red}{95.86}/\color{blue}{89.40} & \color{blue}{95.29}/\color{blue}{89.68} & \color{red}{96.86}/\color{blue}{90.84} & \color{blue}{94.15}/\color{blue}{81.80} & \color{blue}{95.86}/\color{blue}{90.97} \\
    MEAN \cite{MEAN}  & 96.58/89.93 & 96.00/89.49 & 95.15/88.77 & 94.44/87.44 & 93.58/86.91 & 94.44/87.44 & 93.72/86.91 & 96.29/89.87 & 92.87/79.66 & 95.44/86.05 \\
    Ours  & \color{blue}{97.15}/\color{red}{96.21} & \color{red}{97.00}/\color{red}{95.73} & \color{red}{97.15}/\color{red}{94.54} & \color{red}{97.43}/\color{red}{94.77} & \color{red}{96.58}/\color{red}{93.74} & \color{blue}{95.72}/\color{red}{93.39} & \color{red}{96.58}/\color{red}{93.84} & \color{red}{96.86}/\color{red}{95.75} & \color{red}{95.86}/\color{red}{90.97} & \color{red}{96.86}/\color{red}{95.35} \\
    \bottomrule
    \end{tabular}%
    }
  \label{tab:university_weather}%
\end{table*}%

\begin{table*}[ht]
  \centering
  \caption{Comparison results between the proposed method and state-of-the-art approaches under cross-domain evaluation for the ``drone$\rightarrow$satellite'' task. The best and second-best results are highlighted in \textcolor{red}{\textbf{red}} and \textcolor{blue}{\textbf{blue}}, respectively.}
  \resizebox{\linewidth}{!}{
    \begin{tabular}{lcccccccccc}
    \toprule
    \multicolumn{1}{c}{\multirow{3}[6]{*}{Model}} & \multirow{3}[6]{*}{Parameters(M)} & \multirow{3}[6]{*}{GFLOPs} & \multicolumn{8}{c}{Drone$\rightarrow$Satellite} \\
    % \multirow{3}[6]{*}{Model} & \multirow{3}[6]{*}{Parameters(M)} & \multirow{3}[6]{*}{GFLOPs} & \multicolumn{8}{c}{Drone$\rightarrow$Satellite} \\
\cmidrule{4-11}          &       &       & \multicolumn{2}{c}{150m} & \multicolumn{2}{c}{200m} & \multicolumn{2}{c}{250m} & \multicolumn{2}{c}{300m} \\
\cmidrule{4-11}          &       &       & R@1   & AP    & R@1   & AP    & R@1   & AP    & R@1   & AP \\
    \midrule
    MCCG \cite{MCCG}  & 56.65  & \color{blue}{51.04} & 57.62  & 62.80  & 66.83  & 71.60  & 74.25  & 78.35  & 82.55  & 85.27  \\
    Sample4Geo \cite{Sample4geo} & 87.57  & 90.24  & 70.05  & 74.93  & 80.68  & 83.90  & 87.35  & 89.72  & 90.03  & 91.91  \\
    DAC \cite{DAC}   & 96.50  & 90.24  & 76.65  & 80.56  & 86.45  & 89.00  & \color{blue}{92.95} & \color{blue}{94.18} & 94.53  & 95.45  \\
    MEAN \cite{MEAN}  & \color{blue}{36.50} & \color{red}{26.18} & \color{blue}{81.73} & \color{blue}{87.72} & \color{blue}{89.05} & \color{blue}{91.00} & 92.13  & 93.60  & \color{blue}{94.63} & \color{blue}{95.76}  \\
    \midrule
    Ours  & \color{red}{10.67} & 86.63  & \color{red}{92.75} & \color{red}{94.05} & \color{red}{95.58} & \color{red}{96.29} & \color{red}{97.00} & \color{red}{97.48} & \color{red}{97.35} & \color{red}{97.80} \\
    \bottomrule
    \end{tabular}%
    }
  \label{tab:Crossdomain_D2S}%
\end{table*}%

\subsection{Comparison with State-of-the-Art Methods}

To validate the effectiveness of the proposed efficient adaptation framework for visual foundation models (BGG) in drone-satellite cross-view geo-localization, we systematically compare it with current state-of-the-art methods. The results for University-1652 are presented in Table~\ref{tab:university}, SUES-200 in Table~\ref{tab:SUE-D2S} and Table~\ref{tab:SUE-S2D}, and Multi-weather University-1652 in Table~\ref{tab:university_weather}.

\textbf{Results on University-1652.}  As shown in Table~\ref{tab:university}, our method achieves state-of-the-art performance under both task settings.  For the drone$\rightarrow$satellite task, we obtain 96.24\%/96.81\% in R@1/AP,  improving over the second-best method DAC \cite{DAC} by 1.57\%/1.31\%, respectively. For the satellite$\rightarrow$drone task, we reach 97.57\%/95.44\% in R@1/AP, 
surpassing DAC by 1.14\%/1.65\%, respectively. Compared with the latest transformer-based method Game4Loc \cite{Game4loc} , our R@1 is higher by 4.92\% and 3.14\% on the two tasks, respectively. This demonstrates that BGG can more effectively bridge the cross-view domain gap and learn robust geometric structural representations. In terms of efficiency, our method requires only 10.67M trainable parameters, which is substantially fewer than DAC (96.50M) and SRLN \cite{SRLN}  (193.03M), while achieving higher accuracy. Moreover, although the lightweight method MEAN \cite{MEAN} has lower GFLOPs (26.18), its accuracy is clearly inferior (e.g., R@1/AP for drone$\rightarrow$satellite is only 93.55\%/94.53\%). 
The above experimental results demonstrate that we have achieved a superior trade-off between performance and model size, making our framework more suitable for CVGL tasks.
% This indicates that we have achieved a more optimal trade-off between performance and model size, making our framework more suitable for CVGL tasks.

\textbf{Results on SUES-200.} As presented in Tables~\ref{tab:SUE-D2S} and~\ref{tab:SUE-S2D}, to evaluate robustness against drastic scale variations, we conduct experiments across four flight altitudes (150m, 200m, 250m, and 300m). For the drone$\rightarrow$satellite task, our method achieves 99.30\%/99.46\% in R@1/AP at the most challenging altitude of 150m, outperforming DAC by 2.50\%/1.92\%, respectively, and it remains the best at 200m and 250m. Although MEAN exhibits slightly higher performance at 300m, our method demonstrates superior overall stability and performance across all altitudes. This trend indicates that our multi-granularity feature enhancement adapter (MFEA) effectively captures scale-invariant geometric features, enabling the model to cope with the large scale shifts in CVGL. For the satellite$\rightarrow$drone task, we obtain the best performance at 150m with 98.75\%/98.22\% in R@1/AP, surpassing MEAN by 1.25\%/3.47\%, respectively.  Meanwhile, although MEAN achieves 100\% R@1 at higher altitudes, it exhibits more pronounced AP fluctuations on challenging low-altitude samples. In contrast, our method maintains both R@1 and AP above 98\% across all altitudes, demonstrating stronger stability and robustness. These results further confirm that MFEA facilitates the pre-trained model in adapting to drastic scale variations and geometric deformations in cross-view localization, while the FASA module effectively captures stable structural information of the images.

\textbf{Results on Multi-weather University-1652.} Real-world scenarios often faces harsh environmental conditions, such as variations in weather and illumination, which significantly degrade the quality of visual features. As shown in Table~\ref{tab:university_weather}, we achieve superior performance across all 10 weather/illumination perturbations under both drone$\rightarrow$satellite and satellite$\rightarrow$drone settings. Specifically, under ``Fog+Snow'' and ``Rain+Snow'' conditions, the R@1 scores for the drone$\rightarrow$satellite task reach 92.63\% and 92.52\%, outperforming DAC by 3.34\% and 2.61\%, respectively. In the ``Over-exposure'' scenario, our R@1/AP is 88.33\%/90.28\%, outperforming DAC by 5.92\%/5.10\%. Similarly, under ``Dark'' conditions, the R@1/AP reaches 96.04\%/96.78\%, surpassing DAC by 5.04\%/4.24\%, respectively. This corroborates that our frequency-aware structural aggregation (FASA) module effectively suppresses high-frequency noise caused by environmental variations while preserving stable geometric structural features, such as buildings. Moreover, for the satellite$\rightarrow$drone task, our AP remains above 90\% across all adverse conditions,  indicating consistently strong adaptability to environmental variations.

Overall, the results on the three benchmark datasets validate the advantages of our method from three perspectives: robustness to geometric variations, adaptability to environmental changes, and parameter efficiency. Our approach models multi-scale spatial relations from local textures to broader spatial context via MFEA, and further mines deep, stable geometric structural information with FASA. Together, our method enable efficient transfer of vision foundation model DINOv3 to CVGL, while maintaining the consistency and robustness of cross-view features.

\begin{table*}[ht]
  \centering
  \caption{Comparison results between the proposed method and state-of-the-art approaches under cross-domain evaluation for the ``satellite $\rightarrow$drone'' task. The best and second-best results are highlighted in \textcolor{red}{\textbf{red}} and \textcolor{blue}{\textbf{blue}}, respectively.}
  \resizebox{\linewidth}{!}{
    \begin{tabular}{lcccccccccc}
    \toprule
    \multicolumn{1}{c}{\multirow{3}[6]{*}{Model}} & \multirow{3}[6]{*}{Parameters(M)} & \multirow{3}[6]{*}{GFLOPs} & \multicolumn{8}{c}{Satellite$\rightarrow$Drone} \\
    % \multirow{3}[6]{*}{Model} & \multirow{3}[6]{*}{Parameters(M)} & \multirow{3}[6]{*}{GFLOPs} & \multicolumn{8}{c}{Satellite$\rightarrow$Drone} \\
\cmidrule{4-11}          &       &       & \multicolumn{2}{c}{150m} & \multicolumn{2}{c}{200m} & \multicolumn{2}{c}{250m} & \multicolumn{2}{c}{300m} \\
\cmidrule{4-11}          &       &       & R@1   & AP    & R@1   & AP    & R@1   & AP    & R@1   & AP \\
    \midrule
    MCCG \cite{MCCG}  & 56.65  & \color{blue}{51.04} & 61.25  & 53.51  & 82.50  & 67.06  & 81.25  & 74.99  & 87.50  & 80.20  \\
    Sample4Geo \cite{Sample4geo} & 87.57  & 90.24  & 83.75  & 73.83  & \color{blue}{91.25} & 83.42  & 93.75  & 89.07  & \color{blue}{93.75}  & 90.66  \\
    DAC \cite{DAC}   & 96.50  & 90.24  & 87.50  & 79.87  & \color{red}{96.25} & 88.98  & \color{blue}{95.00} & \color{blue}{92.81} & \color{red}{96.25} & 94.00  \\
    MEAN \cite{MEAN}  & \color{blue}{36.50} & \color{red}{26.18} & \color{blue}{91.25} & \color{blue}{81.50} & \color{red}{96.25} & \color{blue}{89.55} & \color{blue}{95.00} & 92.36  & \color{red}{96.25} & \color{blue}{94.32}  \\
    \midrule
    Ours  & \color{red}{10.67} & 86.63  & \color{red}{93.75} & \color{red}{88.46} & \color{red}{96.25} & \color{red}{93.47} & \color{red}{96.25} & \color{red}{96.20} & \color{red}{96.25} & \color{red}{97.01} \\
    \bottomrule
    \end{tabular}%
    }
  \label{tab:crossdomain_S2D}%
\end{table*}%

% 跨域泛化性能对比
\subsection{Cross-Domain Generalization Performance}

In cross-view geo-localization tasks, the generalization capability of a model and its robustness to domain shifts are pivotal indicators for assessing model competence. To evaluate cross-domain generalization, we train the model on University-1652 and directly test it on SUES-200 without any fine-tuning. Since the two datasets differ substantially in scene layouts, viewpoint variations, and scale distributions, this setting provides a stricter assessment of robustness under out-of-distribution (OOD) conditions.

As shown in Table~\ref{tab:Crossdomain_D2S}, our method consistently outperforms existing state-of-the-art methods across all altitude settings in the drone$\rightarrow$satellite task, with the advantage being particularly pronounced in low-altitude scenarios. At the most challenging altitude of 150m, where scale shifts are more severe, existing methods exhibit a clear performance degradation. For instance, the lightweight method MEAN achieves 81.73\%/87.72\% in R@1/AP, and the large-parameter model DAC achieves only 76.65\%/80.56\%. In contrast, our model reaches 92.75\%/94.05\%, representing improvements of 11.02\%/6.33\% over MEAN and 16.10\%/13.49\% over DAC, respectively. These results strongly support the effectiveness of our multi-granularity feature enhancement adapter (MFEA). 
By using multi-level dilated convolutions to model multi-scale spatial relations from local textures to broader spatial context, MFEA encourages scale-invariant representations and enables effective generalization to unseen flight altitudes.

Similarly, as shown in Table~\ref{tab:crossdomain_S2D}, for the satellite$\rightarrow$drone localization task, our method consistently achieves the best performance in terms of both R@1 and AP. In particular, at 150m, we obtain an AP of 88.46\%, which surpasses the second-best method MEAN (81.50\%) by 6.96\%. The higher AP indicates that our model not only successfully retrieves correct matches but also significantly mitigates false positives within the candidate lists, thereby elevating the overall ranking quality. It further confirms the stability and robustness of our approach under cross-domain settings.

Moreover, compared with existing state-of-the-art models such as MEAN and DAC, our method delivers superior generalization with only 10.67M trainable parameters. By efficiently adapting the frozen DINOv3 backbone, we mitigate catastrophic forgetting of general visual knowledge while enhancing the model's perception of scale variations and geometric deformations specific to CVGL. As a result, our approach provides a CVGL solution that is both accurate and efficient under cross-domain scenarios.

% Table generated by Excel2LaTeX from sheet 'Ablation-module'
\begin{table}[t]
  \centering
  \caption{Impact of each component on the performance of the proposed method. ``Frozen'' denotes the frozen pre-trained DINOv3 model. The best results are highlighted in \textcolor{red}{\textbf{red}}.}
  \resizebox{\linewidth}{!}{
    \begin{tabular}{lcccc}
    \toprule
    \multicolumn{1}{c}{\multirow{2}[4]{*}{Model}} & \multicolumn{2}{c}{Drone$\rightarrow$Satellite} & \multicolumn{2}{c}{Satellite$\rightarrow$Drone} \\
\cmidrule{2-5}          & R@1   & AP    & R@1   & AP \\
    \midrule
    Frozen & 38.05  & 43.90  & 29.24  & 17.68  \\
    Frozen+FASA & 84.91  & 87.42  & 91.30  & 83.84  \\
    Frozen+MFEA & 95.64  & 96.35  & 96.86  & 95.33  \\
    Ours  & \color{red}{96.24} & \color{red}{96.81} & \color{red}{97.57} & \color{red}{95.44} \\
    \bottomrule
    \end{tabular}%
    }
  \label{tab:Ablation-module}%
\end{table}%

% Table generated by Excel2LaTeX from sheet 'Ablation-patchtoken'
\begin{table}[t]
  \centering
  \caption{Ablation study on feature descriptor aggregation strategies. ``+'' denotes the channel concatenation operation. ``Mean'', ``Max'', and ``GeM'' represent average, maximum, and generalized mean pooling \cite{GeM} on patch tokens, respectively. NetVLAD is a local feature aggregation method proposed in \cite{NetVLAD}. FASA denotes our proposed frequency-aware structural aggregation module. The best and second-best results are highlighted in \textcolor{red}{\textbf{red}} and \textcolor{blue}{\textbf{blue}}, respectively.}
  \resizebox{\linewidth}{!}{
    \begin{tabular}{lcccc}
    \toprule
    \multicolumn{1}{c}{\multirow{2}[4]{*}{Model}} & \multicolumn{2}{c}{Drone$\rightarrow$Satellite} & \multicolumn{2}{c}{Satellite$\rightarrow$Drone} \\
\cmidrule{2-5}          & R@1   & AP    & R@1   & AP \\
    \midrule
    \texttt{[CLS]}   & 95.64  & 96.35  & \color{blue}{96.86} & 95.33  \\
    \texttt{[CLS]}+Mean & 95.44  & 96.28  & 96.43  & 95.18  \\
    \texttt{[CLS]}+Max & 95.55  & 96.28  & 96.72  & 95.07  \\
    \texttt{[CLS]}+NetVLAD & 95.71  & 96.41  & 96.71  & \color{blue}{95.32} \\
    \texttt{[CLS]}+GeM & \color{blue}{95.91} & \color{blue}{96.59} & 96.72  & 95.16  \\
    \midrule
    \texttt{[CLS]}+FASA & \color{red}{96.24} & \color{red}{96.81} & \color{red}{97.57} & \color{red}{95.44} \\
    \bottomrule
    \end{tabular}%
    }
  \label{tab:Ablation-patchtoken}%
\end{table}%

\begin{figure*}[htbp]  
    \centering
    \includegraphics[width=\textwidth]{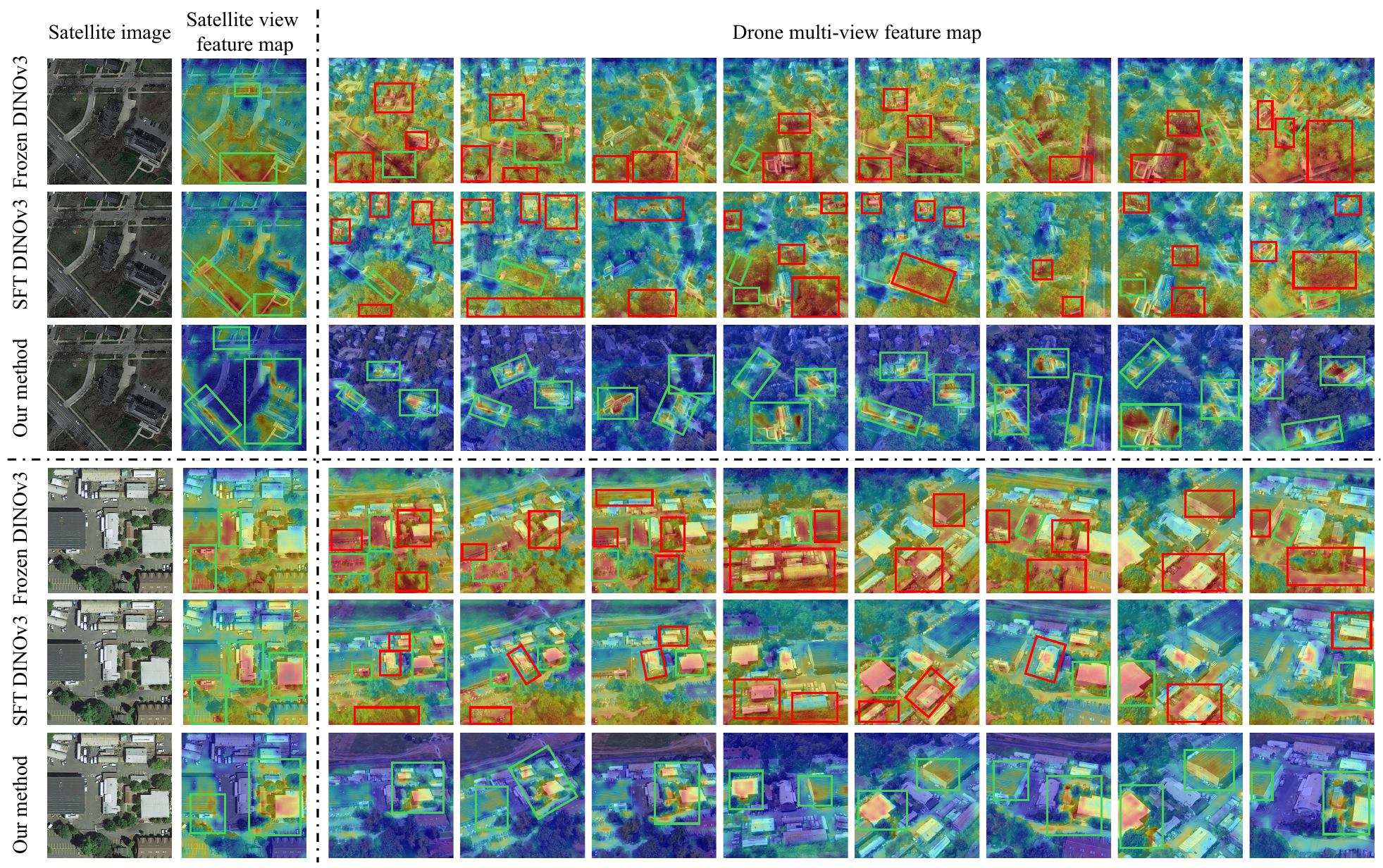}
    \caption{Visualization of cross-view feature maps on University-1652 for the pre-trained foundation model (DINOv3) under the Frozen and full-parameter fine-tuning (SFT) settings, as well as the proposed method. Each group comprises an original satellite image, its corresponding satellite view feature map, and the matching multi-view (varying scales/viewpoints) drone feature maps. The heatmap highlights the regions attended to by the model. For the drone multi-view feature maps, green boxes indicate high-response regions that are consistent with the satellite view salient regions, whereas red boxes denote high-response regions that are inconsistent with the satellite view attention.}
   \label{fig:visual_analyse}
\end{figure*}

% 消融实验
\subsection{Ablation Studies}

In the ablation study, we systematically evaluate the contribution of each component within the proposed method on the University-1652 dataset and further analyze the impact of different feature descriptor aggregation strategies on localization performance.

As shown in Table~\ref{tab:Ablation-module}, we quantify the impact of the multi-granularity feature enhancement adapter (MFEA) and the frequency-aware structural aggregation (FASA) module on overall performance. The ``Frozen'' baseline, which directly utilizes pre-trained features without any task-specific adaptation, performs poorly. For example, on the drone$\rightarrow$satellite task, it achieves only 38.05\%/43.90\% in R@1/AP. This result indicates a substantial domain gap between the pre-training objective and cross-view geo-localization, which is highly sensitive to geometric variations. Integrating the MFEA into the frozen backbone delivers a substantial performance boost, elevating R@1/AP to 96.24\%/96.81\% (an improvement of 58.19\%/52.91\%). This significant gain suggests that by modeling multi-scale spatial relationships ranging from local textures to spatial contexts via multi-level dilated convolutions, the MFEA enables the pre-trained foundation model to effectively cope with inherent viewpoint discrepancies and geometric deformations in CVGL. Meanwhile, incorporating the FASA module alone also significantly improves R@1/AP to 84.91\%/87.42\%, suggesting that the structural aggregation of local features effectively compensates for the deficiency of global representations in capturing spatial structural details. Finally, combining both modules yields the best performance, reaching 96.24\%/96.81\%. This demonstrates the strong complementarity between MFEA and FASA. Specifically, MFEA better adapts the pre-trained DINOv3 to the geometric and scale variations in CVGL, while FASA optimizes the final global representation by highlighting stable local structural information.

To validate the rationality of the FASA module, we compare it against mainstream feature aggregation methods, encompassing both simple pooling (Mean, Max) and more advanced aggregation techniques (NetVLAD, GeM). All methods are applied to the patch tokens output by the adapted backbone, and the aggregated feature is concatenated with the \texttt{[CLS]} token along the channel dimension.

As shown in Table~\ref{tab:Ablation-patchtoken}, simple aggregation methods (e.g., Mean and Max pooling) do not provide meaningful gains over using the \texttt{[CLS]} token alone, and may even cause slight degradation. For example, on the drone$\rightarrow$satellite task, mean pooling yields 95.44\%/96.28\% in R@1/AP,  representing a decrease of 0.20\%/0.07\% compared to the \texttt{[CLS]}-only baseline. More advanced aggregation methods (e.g., GeM pooling) achieve competitive results, reaching 95.91\%/96.59\% in R@1/AP. However, such methods are less stable overall; on the satellite$\rightarrow$drone task, their R@1 is even inferior to the \texttt{[CLS]}-only baseline. In contrast, our FASA module consistently outperforms all comparative methods across both drone$\rightarrow$satellite and satellite$\rightarrow$drone tasks, achieving the highest 96.24\%/96.81\% and 97.57\%/95.44\% in R@1/AP, respectively. The superiority of FASA stems from its frequency-aware and structural information enhancement. Under substantial cross-view discrepancies, FASA utilizes frequency-domain modulation, feature gated mixing and adaptive weighting mechanisms to suppress unstable interference while emphasizing structurally consistent regions (e.g., roads and building boundaries). Consequently, it produces more robust descriptors and improves adaptability to complex geometric variations in cross-view geo-localization.

\begin{figure*}[htbp]  
    \centering
    \includegraphics[width=\textwidth]{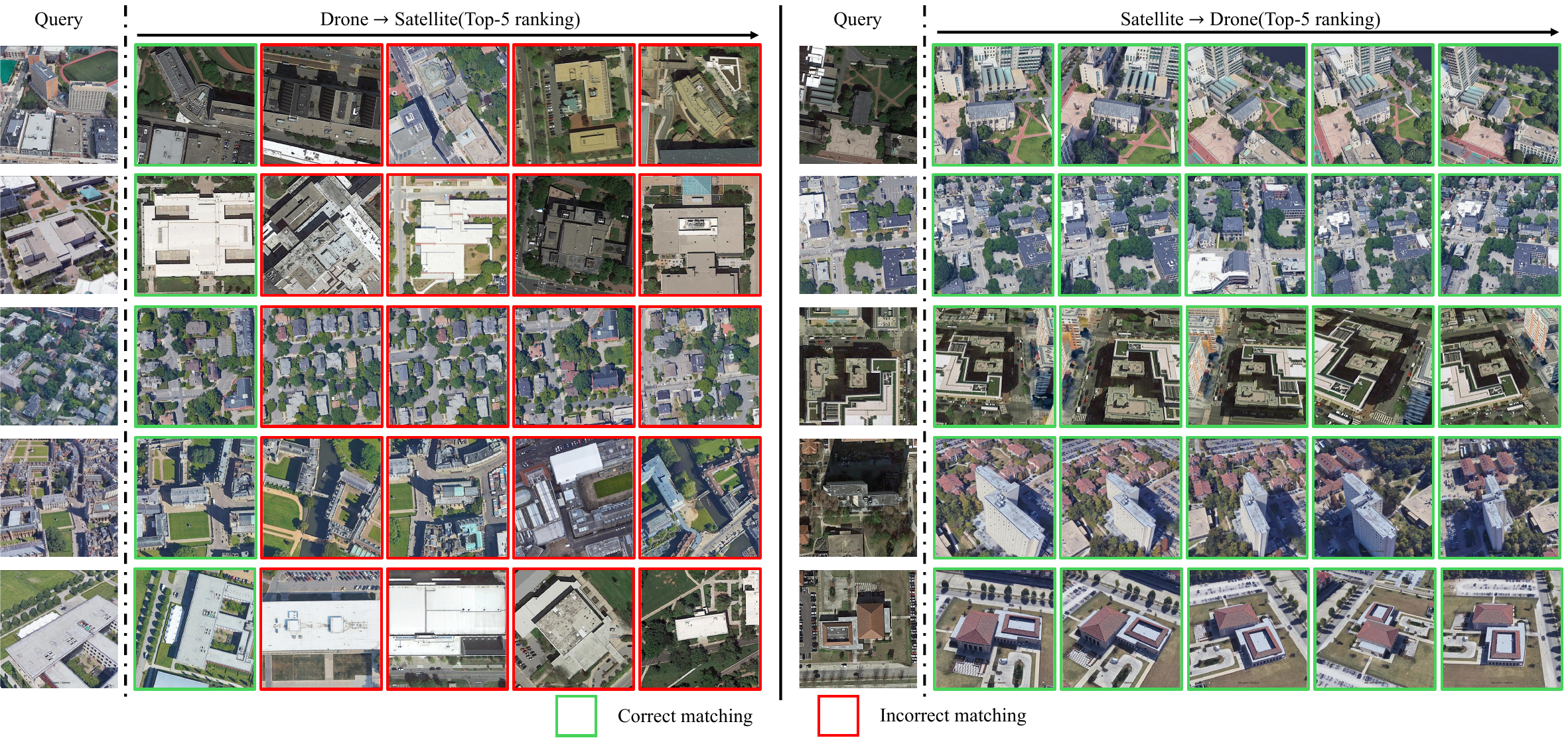}
    \caption{Top-5 retrieval results of the proposed method on the University-1652 dataset.}
   \label{fig:retrival_university}
\end{figure*}

\begin{figure*}[htbp]  
    \centering
    \includegraphics[width=\textwidth]{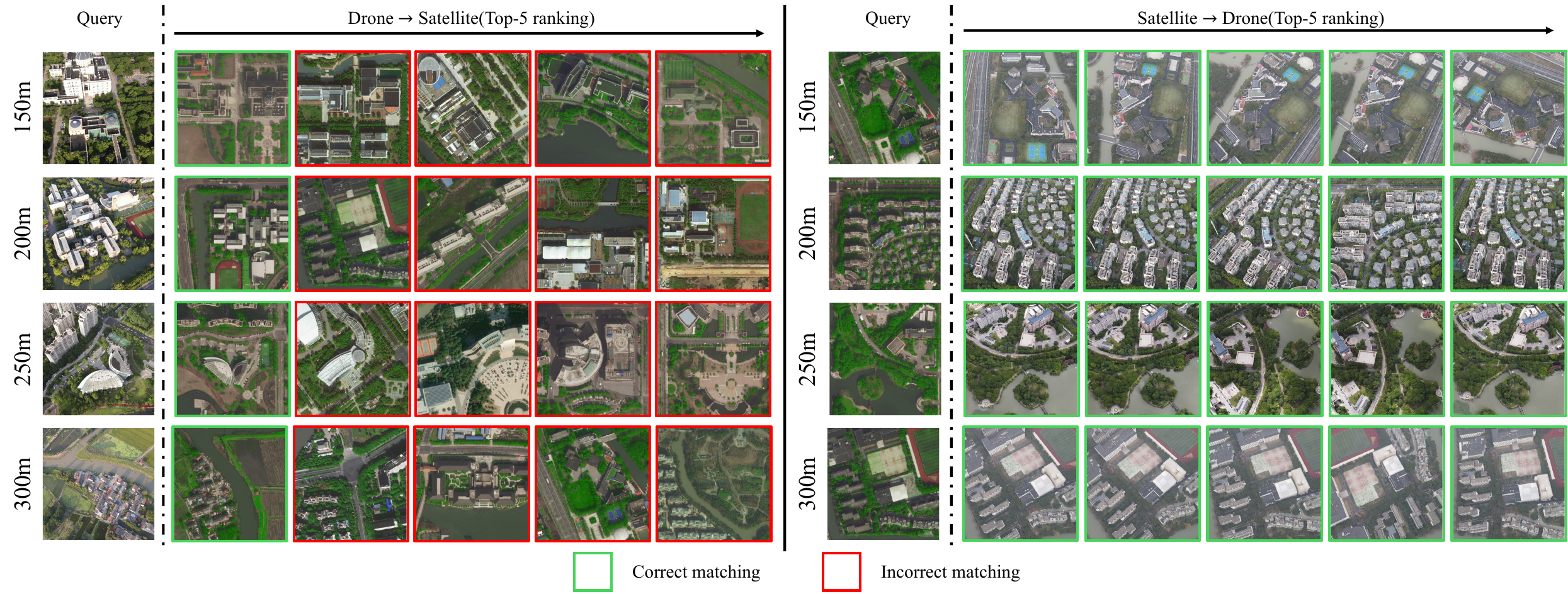}
    \caption{Top-5 retrieval results of the proposed method on the SUES-200 dataset.}
   \label{fig:retrival_sue}
\end{figure*}

% 可视化分析
\subsection{Visualization Analysis}

To visually demonstrate the representation consistency of the proposed method under cross-view conditions, we compare the feature maps of the Frozen DINOv3, SFT DINOv3, and our method on the University-1652 dataset. Each group comprises an original satellite image, the corresponding satellite view feature map, and drone feature maps of the same location under 8 different scales/viewpoints. High response regions in the feature maps indicate the model's attended areas. Specifically, green boxes denote attended regions that align with high response areas in the satellite view, while red boxes indicate high response regions that are inconsistent with the satellite view. The visualization shows that our model exhibits superior cross-view attentional consistency.

As shown in the first and fourth rows of Fig.~\ref{fig:visual_analyse}, the frozen DINOv3 backbone struggles to effectively localize targets; its attention is relatively dispersed and susceptible to interference from high-frequency background noise (e.g., vegetation textures). Furthermore, the salient regions in the satellite view are difficult to correspond to those in the multi-view drone feature maps,  indicating weak cross-view consistency.  This suggests that without task-specific adaptation, the general semantic representations learned during pretraining are highly sensitive to view-inconsistent textures between satellite and drone images, and thus fail to achieve reliable cross-view alignment. SFT DINOv3 (as shown in the second and fifth rows) improves localization capabilities to some extent, 
and exhibits partially consistent cross-view responses (more green boxes). However, many inconsistent high-response regions remain. This suggests that direct full-parameter fine-tuning is not sufficiently robust to geometric deformations, and it is difficult to learn geometry-invariant structural representations.

In contrast, our method (as shown in the third and sixth rows) exhibits significantly stronger cross-view attentional consistency. In both satellite and drone views, the model precisely focuses on structurally stable regions across views, such as building edges and road intersections. Even under significant scale variations and viewpoint transformations, the model maintains sustained attention on key structural regions without producing widespread responses in irrelevant regions. This phenomenon verifies that the MFEA enhances the model's adaptability to cross-view scale variations and geometric distortions by modeling multi-scale spatial relationships ranging from local textures to spatial contexts via multi-level dilated convolutions. Meanwhile, the FASA module further emphasizes cross-view consistent structural information, improving the robustness and cross-view consistency of the final descriptor. Together, they enable efficient adaptation of the vision foundation model DINOv3 for CVGL.

% 检索结果分析
\subsection{Retrieval Result Analysis}

To further demonstrate the effectiveness of the proposed method, Figs.~\ref{fig:retrival_university} and~\ref{fig:retrival_sue} visualize the Top-5 retrieval results for representative query samples from the University-1652 and SUES-200 datasets, respectively. In these figures, Green borders denote correctly retrieved matches, while red borders indicate incorrect matches.

For the drone$\rightarrow$satellite localization task, each drone query image corresponds to a unique matched satellite image in the gallery. As illustrated in Fig.~\ref{fig:retrival_university}, our method successfully identifies the correct satellite image at the Top-1 rank for all examples, demonstrating high accuracy and stability in the one-to-one matching setting. For the satellite$\rightarrow$drone localization task, each satellite query corresponds to multiple drone images captured at the same location. Our method accurately retrieves matching drone images within the Top-5 candidate lists. This indicates that the model effectively captures cross-view stable structural information, such as road topology and building layouts, thereby maintaining robust alignment capabilities despite cross-view scale variations and geometric discrepancies.

Moreover, to validate robustness under cross-view multi-scale conditions, we visualize retrieval results at different flight altitudes (150m, 200m, 250m, and 300m) on the SUES-200 dataset. As shown in Fig.~\ref{fig:retrival_sue}, for the drone$\rightarrow$satellite task, despite the increasing field-of-view changes and scale discrepancies induced by altitude variations, our method still consistently retrieves the unique correct match. Similarly, for the drone$\rightarrow$satellite task, the Top-5 candidate lists consistently contain correct matches across all altitudes.  These results demonstrate that our method achieves precise localization even under extreme scale variations and viewpoint differences.

\section{Conclusion}
This paper proposes BGG, a parameter-efficient adaptation framework based on the VFM for cross-view geo-localization, which aims to bridge the geometric gap between cross-view images. Specifically, BGG employs the multi-granularity feature enhancement adapter (MFEA) for improving feature robustness to scale changes and geometric variations between cross-view images. Additionally, BGG utilizes the frequency-aware structural aggregation (FASA) module to extract discriminative local structural information by patch token modulation in the frequency domain and adaptive aggregation. Finally, BGG concatenates the local features with the global \texttt{[CLS]} token for more precise cross-view matching and localization. Extensive experiments demonstrate that BGG not only surpasses state-of-the-art methods in localization accuracy but also maintains strong generalization across various cross-view scenarios with small training costs. In future work, we will further explore the potential of VFMs for higher-precision visual localization tasks.

\bibliographystyle{IEEEtran}
\bibliography{references}

\vfill

\end{document}